\documentclass{bmvc2k}

\usepackage{graphicx}
\usepackage{amsmath}
\usepackage{fixltx2e}
\graphicspath{{images/}}

\title{2sRanking-CNN: A 2-stage ranking-CNN for diagnosis of glaucoma from fundus images using CAM-extracted ROI as an intermediate input}

\addauthor{Tae Joon Jun}{taejoon89@kaist.ac.kr}{1}
\addauthor{Dohyeun Kim}{stalker7@kaist.ac.kr}{1}
\addauthor{Hoang Minh Nguyen}{minhhoang@kaist.ac.kr}{1}
\addauthor{Daeyoung Kim}{kimd@kaist.ac.kr}{1}
\addauthor{Youngsub Eom}{hippotate@hanmail.net}{2}

\addinstitution{
 School of Computing \\
 Korea Advanced Institute of Science and Technology \\
 Daejeon, Republic of Korea.
}
\addinstitution{
 Department of Ophthalmology\\
 Korea University College of Medicine\\
 Seoul, Republic of Korea.
}

\runninghead{Student, Prof, Collaborator}{BMVC Author Guidelines}

\begin{document}

\maketitle

%-------------------------------------------------------------------------
\begin{abstract}
Glaucoma is a disease in which the optic nerve is chronically damaged by the elevation of the intra-ocular pressure, resulting in visual field defect. Therefore, it is important to monitor and treat suspected patients before they are confirmed with glaucoma. In this paper, we propose a 2-stage ranking-CNN that classifies fundus images as normal, suspicious, and glaucoma. Furthermore, we propose a method of using the class activation map as a mask filter and combining it with the original fundus image as an intermediate input. Our results have improved the average accuracy by about 10\% over the existing 3-class CNN and ranking-CNN, and especially improved the sensitivity of suspicious class by more than 20\% over 3-class CNN. In addition, the extracted ROI was also found to overlap with the diagnostic criteria of the physician. The method we propose is expected to be efficiently applied to any medical data where there is a suspicious condition between normal and disease.
\end{abstract}
%-------------------------------------------------------------------------
\section{Introduction}
\label{sec:introduction}
Glaucoma is an eye disease that causes narrowed vision and eventually leads to blindness, which is caused by various reasons such as elevated intra-ocular pressure (IOP) or blood circulation disorder. Once glaucoma is diagnosed, it needs constant management for a lifetime, and the damaged vision is not restored. Therefore, early detection and treatment of glaucoma is the best prevention, but the optic nerve damage caused by glaucoma gradually develops, and when symptoms appear, the disease progresses considerably. In addition, since it is not easy to confirm glaucoma early, various tests including IOP measurement, optic nerve head examination, and anterior chamber angle examination are conducted and the results are combined to determine the existence of glaucoma.

% Literature 정리 후 추가 필요함
Therefore, there are several previous studies to classify normal and glaucoma in fundus image through machine learning and to play a supporting role in physician's glaucoma diagnosis criteria. Chen performed a classification of normal and glaucoma using a convolutional neural network in \cite{chen2015glaucoma}. Chen designed the AlexNet-style \cite{krizhevsky2012imagenet} CNN, evaluated with the ORIGA \cite{zhang2010origa} and SCES \cite{sng2012determinants} fundus image dataset, and obtained 0.831 and 0.887 area under the curve (AUC), respectively. Chen's study is significant in that it classifies glaucoma using CNN, but classifies only normal and glaucoma classes and does not show good classification performance. Li proposed a model combining CNN and SVM to diagnose glaucoma focusing on the disk/cup region of interests (ROI) and obtained a 0.838 AUC in \cite{li2016integrating}. Li's work, however, has the same limitations as Chen's work, and at the same time, did not directly extract the disk/cup ROI, but instead used the ROI that was manually labeled in the ORIGA dataset. Khali conducted a review of several machine learning techniques for glaucoma detection in \cite{khalil2014review}. Various machine learning techniques have been compared such as decision tree, fuzzy logic, K-nearest neighbor, support vector machine, and Naive Bayes.

However, none of the studies described above take into account the intermediate state of normal and glaucoma, and classification performance is not excellent. Moreover, since this intermediate class is a continuous state between normal and glaucoma, classification using ranking-CNN \cite{chen2017using} seems appropriate. Therefore, We propose a 2-stage ranking-CNN (2sRanking-CNN) classifying fundus images labeled normal, suspicious, and glaucoma. 2sRanking-CNN uses the class activation map (CAM) \cite{zhou2016learning} of the 1st-stage model, which is trained lightly by train-set and validation-set, as a mask filter for ROI extraction. Since the ranking-CNN consists of binary classification models, the CAM of the suspicious class uses the average value of the CAM of the models constituting the ranking-CNN. The extracted CAM is integrated with the original fundus image and used as the input of the 2nd-stage model. Then, the classification result of the 2nd-stage model is used as the final prediction. Our 2sRanking-CNN compares accuracy with single-stage ranking-CNN and 3-class CNN that classify normal, suspicious, and glaucoma simultaneously. As a result, 2sRanking-CNN achieved an average accuracy of 96.46\%, specificity of 96\%, sensitivity for suspicious of 97.56\% and sensitivity for glaucoma of 95.18\%. Based on average accuracy, 2sRanking-CNN is 9.61\% and 10.6\% higher than ranking-CNN and 3-class CNN, respectively, and surprisingly 14.63\% and 24.39\% higher for sensitivity for suspicious. In addition, the highlighted area of CAM we obtained as a result of the 1st-stage model included the reference area where the ophthalmologist diagnoses glaucoma in a given fundus image. Consequently, we expect that our 2sRanking-CNN can be similarly applied to any medical imaging data with an intermediate state between normal and disease.
%-------------------------------------------------------------------------
\section{Methods}
\label{sec:methods}
\subsection{Data acquisition}
%데이터를 얻은 과정 (몇 명의 환자, 어떤 환자 등) 그리고 정상/의심/녹내장을 label한 기준
This study included 1022 fundus images from 301 consecutive patients (582 eyes) who underwent fundus imaging with a non-mydriatic fundus camera (TRC-NW8; Topcon, Oakland, NJ, USA), between January 2016 and August 2017. During the study period, patient electronic medical records and fundus imaging were reviewed to determine the presence of glaucoma by the glaucoma specialist. Based on fundus imaging and electronic medical records, 1022 fundus images were divided into three categories; normal, glaucoma suspect (suspicious), and glaucoma. This study adhered to the Declaration of Helsinki and approval for retrospective review of clinical records was obtained from the committee. The patient information was completely anonymized and de-identified prior to analysis. Of the 301 patients, 138 (45.8\%) were men and 163 were women. The mean age ($\pm$ SD) was 59.7 ($\pm$ 15.4) years (range, 19-92 years). There were 291 right eyes (50.0\%) and 291 left eyes. Of the 1022 fundus imaging, 403 (39.4\%) were normal, 208 (20.4\%) were glaucoma suspect, and 381 (37.3\%) were glaucoma. Of these, 992 were used as the fundus image dataset of this study and 30 images with the wrong file format were excluded.

\subsection{2sRanking-CNN}
\begin{figure}[!tbh]
	\centering
	\includegraphics[width=\textwidth]{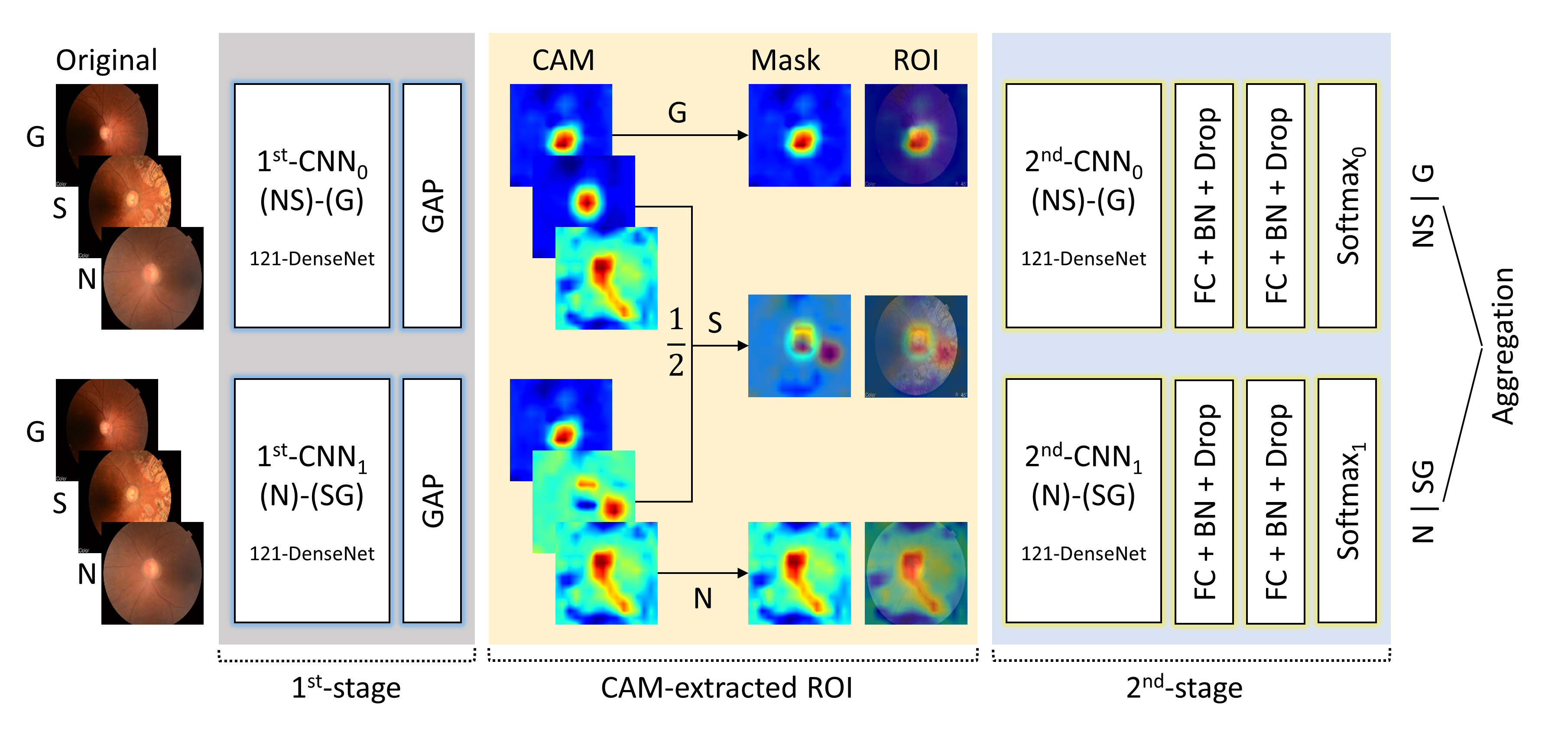}
	\caption{The overall architecture of 2sRanking-CNN}
	\label{overall_archi}
\end{figure}

%전체 architecture를 하나의 그림으로 나타내자
2sRanking-CNN consists of first stage ranking-CNN, steps to extract ROI from CAM (CAM-extracted ROI), and second stage ranking-CNN. 1st-stage ranking-CNN inputs the original fundus image and outputs the CAM mask filter image. Since our fundus image dataset consists of three classes, ranking-CNN consists of two binary classifications. For convenience, the case of grouping normal and suspicious into one class is referred to as (NS)-(G) and grouping suspicious and glaucoma into one class as (N)-(SG). In the CAM-extracted ROI stage, the CAM mask filter image is combined with the original fundus image to become the ROI. The definition of the mask filter used for each class is described in the Section \ref{CAM-ROI}. 2nd-stage ranking-CNN takes the above ROI as an input, trains ranking-CNN once again, and outputs the final prediction value. The overall architecture of 2sRanking-CNN is shown in Figure \ref{overall_archi}
%-------------------------------------------------------------------------
\subsubsection{1st-stage Ranking-CNN}
%Ranking-CNN을 수행하고 CAM을 얻기 전까지의 과정
Before explaining 1st-stage ranking-CNN, let's briefly explain how ranking-CNN works. Ranking-CNN was proposed by Chen in \cite{chen2017using} for age estimation from human face images. For example, considering CNN classifying \textit{N} classes, it is general to classify multi-label classification with \textit{N}-sized softmax layer in final prediction. However, if the class is continuous and the boundaries are ambiguous, general multi-label classification may not work well. Age estimation is a typical example, and diseases with grade can also be an example. Ranking-CNN creates \textit{N}-1 small CNN models for class classification, and each model performs binary classification with one class as a reference point. For example, when predicting the age from 10 to 50 years old, the first CNN model is based on the age of 11, and the tenth model is a binary classification based on 20 years old. As a result, \textit{N}-1 binary predictions are obtained for the \textit{N} classes, and the classification of the class is the number of true values. Similarly, in the age estimation example, 10-year-old has zero true value and 20-year-old has 10 true values. We introduced ranking-CNN in the glaucoma diagnosis because we determined that ranking-CNN could be applied efficiently because our class is also continuous and bounded in a similar way as predicting age. In addition, 2sRanking-CNN is proposed to efficiently classify ambiguous between classes rather than simply applying ranking-CNN.

Our fundus image dataset consists of three classes: normal, suspicious, and glaucoma, so ranking-CNN is composed of two sub-classifiers. The goal of 1st-stage ranking-CNN is to train two sub-classifiers by train-set and validation-set to obtain CAM as mask filter. To extract the CAM, the layer just before the softmax layer should be a global average pooling (GAP) or a global max pooling, and not a fully-connected (FC) layer. Experiments have shown that GAP is more efficient than GMP in \cite{zhou2016learning}. In addition, a sub-classifier can be a deep CNN like ResNet \cite{he2016deep} or DenseNet \cite{huang2017densely} because unlike the case of age estimation, only two sub-classifiers are needed. In our case, we used 121-layer DenseNet as a sub-classifier. After training a certain degree of epochs, in our case 20 epochs, we aggregate the predictions of two sub-classifiers to predict the final class. The important point is that when extracting the CAM, it should be based on a predicted class rather than an actual class. If the predicted class is wrong in the 1st-stage, it can be modified by comparing it with other ROIs of the same class in 2nd-stage. In addition, because the test set does not know the actual class in 1st-stage, the test set can not have an ROI if it is extracted based on the actual class. As result, 1st-stage ranking-CNN outputs 3 CAMs for each sub-classifier, resulting in a total of 6 CAMs. How each CAM is used as a mask filter image is discussed in the next section.
%-------------------------------------------------------------------------
\subsubsection{CAM-extracted ROI}
\label{CAM-ROI}
%sub-model 별로 얻은 CAM 들 중에서 어떤 것들을 intermediate input으로 선택했고, 그 이유에 대한 근거
%original하고 합쳐져서 input이 되는 과정까지 보여준다
The CAMs obtained from the 1st-stage are used as a mask filter image and combined with the original fundus image to generate ROI. CAM is a method that Zhou introduces in \cite{zhou2016learning} and performs the inner product of the feature maps immediately before the GAP layer and the weights of the softmax layer and displays them in image form. As a result, when we calculate the probability to classify into class C, we multiply each weight to feature map and sum it up, and it is possible to visualize by what criteria the model classifies the input image. To add a more formal description of the CAM, let \textit{f}\textsubscript{k}(x,y) be the k-th activation of (x,y) spatial location of an input image. Then, the output value \textit{F}\textsubscript{k} of the GAP layer is $\Sigma$\textsubscript{(x, y)}\textit{f}\textsubscript{k}(x,y). Thus, for given class C, the input \textit{S}\textsubscript{c} for the softmax layer can be expressed as $\Sigma$\textsubscript{k}\textit{w}\textsuperscript{c}\textsubscript{k}\textit{F}\textsubscript{k}, where \textit{w}\textsuperscript{c}\textsubscript{k} represents the weight of class C for unit k. As a result, it can be said that \textit{w}\textsuperscript{c}\textsubscript{k} represents the importance of \textit{F}\textsubscript{k} for a given class C. Substituting \textit{F}\textsubscript{k} = $\Sigma$\textsubscript{(x, y)}\textit{f}\textsubscript{k}(x,y) into \textit{S}\textsubscript{c} yields the following expression:
\begin{equation}\label{(1)}
  \begin{split}
	\textit{S}\textsubscript{c} & = \sum_{k}\textit{w}\textsuperscript{c}\textsubscript{k}\sum_{(x,y)}\textit{f}\textsubscript{k}(x,y)\\
       & = \sum_{(x,y)}\sum_{k}\textit{w}\textsuperscript{c}\textsubscript{k}\textit{f}\textsubscript{k}(x,y)
  \end{split}
\end{equation}
Lets define \textit{M}\textsubscript{c} as the CAM for class C, then \textit{M}\textsubscript{c}(x,y) for (x,y) spatial location is as follows:
\begin{equation}\label{(2)}
	\textit{M}\textsubscript{c} = \sum_{k}\textit{w}\textsuperscript{c}\textsubscript{k}\textit{f}\textsubscript{k}(x,y)
\end{equation}
Therefore, in the above equation, \textit{S}\textsubscript{c} = $\Sigma$\textsubscript{(x, y)}\textit{M}\textsubscript{c}(x,y) and \textit{M}\textsubscript{c}(x,y) refers to the importance of (x,y) spatial location when the given image is classified as class C. Our 2sRanking-CNN is further aimed here to combine the intermediate CAM with the original input image to extract more specific features. In this paper, CAM is applied to ranking-CNN for efficient classification of glaucoma, but it is applicable to general multi-label CNN by replacing the 1st-stage and 2nd-stage sub-classifiers with a single CNN and perform each stage's prediction with multi-label classification. 

\begin{figure}[!tbh]
	\centering
	\includegraphics[width=\textwidth]{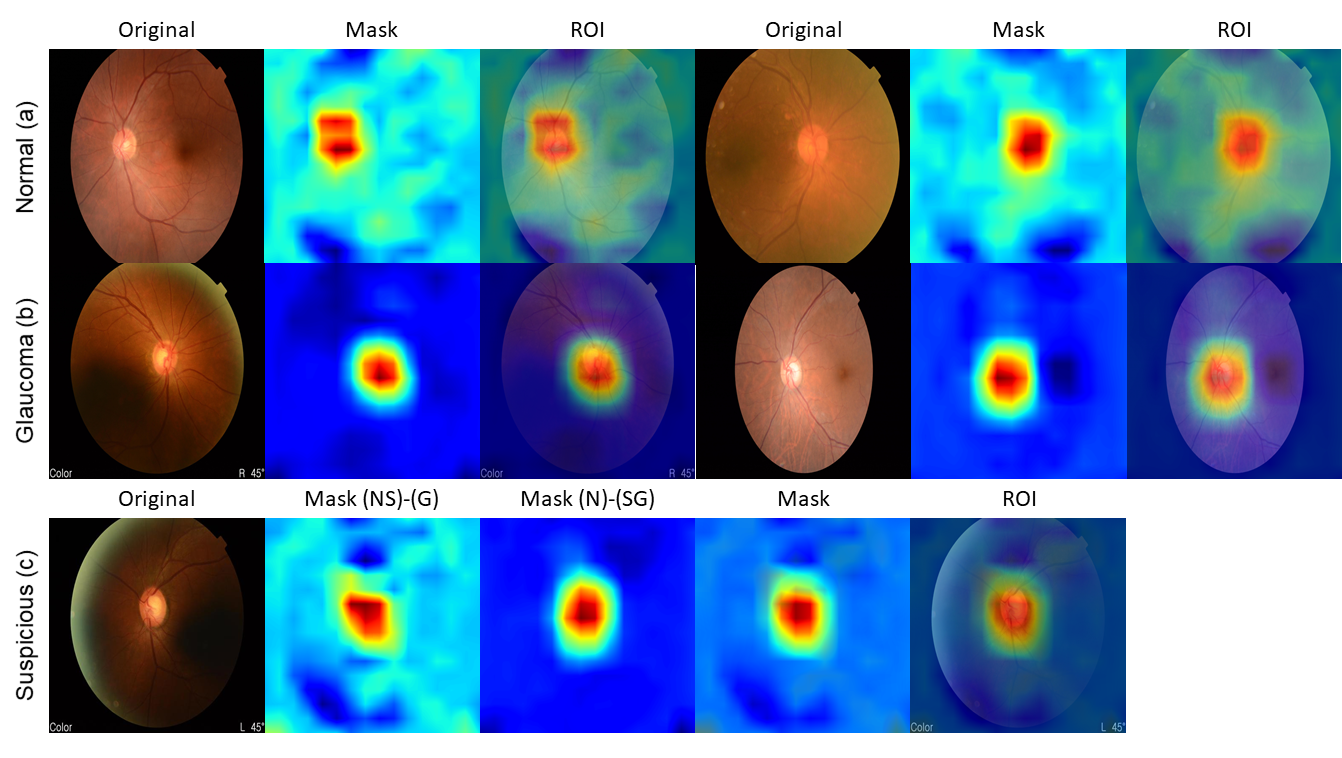}
	\caption{Original fundus images with CAM-extracted mask and ROI images}
	\label{cam_roi}
\end{figure}

The number of CAMs obtained as a result of 1st-stage is 3 per sub-classifier. Here we use the normal class CAM of (N)-(SG) as the mask filter of the input that is normally predicted. In the same way, the mask filter of the input predicted by glaucoma uses the glaucoma class CAM of (NS)-(G). The reason is that the CAM of a class that is classified individually in each group is thought to show more specific characteristics. In case of suspicious class, CAM is close to glaucoma in (N)-(SG) group and close to normal in (NS)-(G) group. Therefore, the mask filter of the suspicious class uses the average of the CAMs of the sub-classifiers. Figure \ref{cam_roi} shows the mask filter images and ROI images generated in the CAM-extracted ROI stage together with the original fundus images. Figure \ref{cam_roi}(a) and (b) show the original, mask filter, and ROI images for normal and glaucoma classes. From the Figure \ref{cam_roi}(a) and (b), it can be seen that CAM is generated by focusing on the disk/cup area in case of glaucoma, while it covers the overall area in addition to the disk/cup area in the normal case. Figure \ref{cam_roi}(c) shows the original, two mask filters, average mask filter, and ROI image of the suspicious class. From the Figure \ref{cam_roi}(c), we can see that the mask filter of the (NS)-(G) group is similar to that of Figure \ref{cam_roi}(a), while the mask filter of the (N)-(SG) group is similar to that of Figure \ref{cam_roi}(b). Therefore, it is reasonable to use the average of the two as a mask filter and obtain an ROI of the suspicious class. Finally, ROI images of each class are used as an intermediate input for 2nd-stage ranking-CNN.
%-------------------------------------------------------------------------
\subsubsection{2nd-stage of 2sRanking-CNN}
%intermediate input으로 FC가 추가된 Ranking-CNN을 학습한다
2nd-stage ranking-CNN trains the model by inputting CAM-extracted ROI images. As in the first stage, sub-classifiers perform binary classification, dividing the dataset into (NS)-(G) and (N)-(SG) groups. 121-layer DenseNet is used as sub-classifiers as same as 1st-stage. Unlike the 1st-stage, the 2nd-stage does not need to extract CAM, so a fully-connected layer can be used. Therefore, we used fully-connected, batch normalization \cite{ioffe2015batch}, and dropout \cite{srivastava2014dropout} layers after the 121-layer DenseNet to strictly prevent overfitting. Finally, the binary prediction of the two sub-classifiers is aggregated to determine the final class.
%-------------------------------------------------------------------------
\section{Results}
\label{sec:results}
\subsection{Experimental setup}
%항상 쓰는 실험 환경 구성
The configuration of 2sRanking-CNN, ranking-CNN, and 3-class CNN for the experiment are as follows. 121-layer DenseNet pre-trained in the ImageNet dataset \cite{deng2009imagenet} was used as a sub-classifier/classifier of each CNN. The ranking-CNN for comparison is the same as the 2nd-stage of 2sRanking-CNN, and the 3-class CNN has the same structure as the sub-classifier of 2sRanking-CNN, but the softmax layer performs 3-class prediction instead of binary classification. RMSprop \cite{tieleman2012lecture} was used as the optimizer function and initial learning rate 0.0001 was set to decrease to 0.9 factor for every epoch. The original fundus image used as the input of the 1st-stage was resized to 512 x 512, and the output CAM was resized from 32 x 32 to 512 x 512 to use as a mask filter. The size of the fully-connected layer of the 2nd-stage was 2048 and the dropout rate was set to 0.5. The ratio of train-set to test-set is 80:20 and 15\% of train-set is set as validation-set. As a result, a total of 992 fundus images are divided into 674, 119, and 199 by train-set, validation-set, and test-set, respectively. 1st-stage ranking-CNN is trained for a total of 20 epochs and outputs CAM based on when the validation loss is the smallest. Similarly, 2nd-stage ranking-CNN is trained for 50 epochs and predicts the final class when the validation loss is minimized. In both 1st-stage and 2nd-stage, we performed image augmentation to prevent overfitting. We zoom-in and zoom-out images at a random rate within 12.5\% and randomly flipped the image horizontally. However, random cropping was not performed because the fundus image itself was photographed to include the entire fundus, there was a concern about the loss of the image features.

The software and hardware environment for the experiment are as follows. We tested on a 32GB server with two NVIDIA Titan X GPUs and an Intel Core\textsuperscript{TM} i7-6700K CPU. The operating system is Ubuntu 16.04, and the development of the CNN model uses Python-based machine learning libraries including Keras, Scikit-learn \cite{pedregosa2011scikit}, and TensorFlow \cite{abadi2016tensorflow}.
%-------------------------------------------------------------------------
\subsection{Evaluation results}
\begin{figure}[!tbh]
	\centering
	\includegraphics[width=\textwidth]{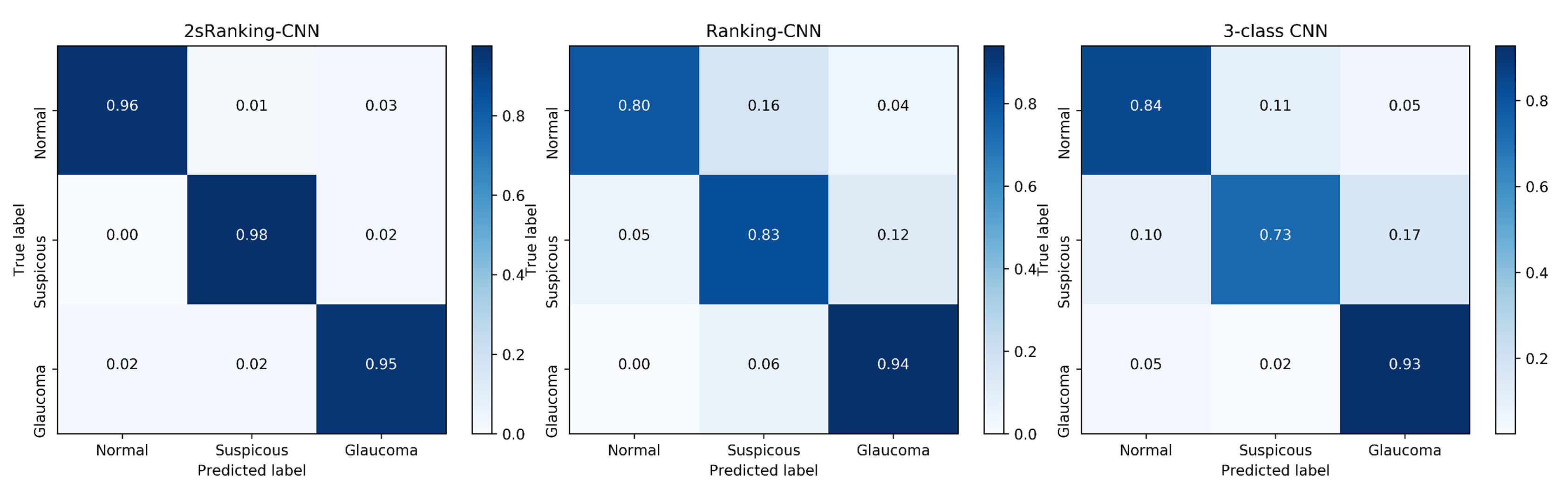}
	\caption{Confusion matrix of 2sRanking-CNN, ranking-CNN, and 3-class CNN}
	\label{confusion_matrix}
\end{figure}

The evaluation of the glaucoma classification was based on the following four metrics: average accuracy (\textit{Acc}), specificity (\textit{Sp}), sensitivity for suspicious (\textit{Se}\textsuperscript{S}), and sensitivity for glaucoma (\textit{Se}\textsuperscript{G}). Average accuracy means a correctly predicted percentage of the total data. Specificity, also known as the true negative rate, measures the percentage of negatives that are correctly identified as normal. Sensitivity, also known as the true positive rate or recall, measures the percentage of positives that are correctly identified as suspicious or glaucoma. Table \ref{result_summary} summarizes the performance evaluation results of 2sRanking-CNN, ranking-CNN, and 3-class CNN based on evaluation metrics. For a more specific evaluation, confusion matrix for each method is presented in Figure \ref{confusion_matrix}.
\begin{table}[!tbh]
  \begin{center}
    \begin{tabular}{|l|c|c|c|c|}
     \hline
      Method & \textit{Acc}(\%) & \textit{Sp}(\%) & \textit{Se}\textsuperscript{S}(\%) & \textit{Se}\textsuperscript{G}(\%) \\
     \hline\hline
     2sRanking-CNN & 96.46 & 96.00 & 97.56 & 95.18\\
     Ranking-CNN & 86.87 & 80.00 & 82.93 & 93.98\\
     3-class CNN & 85.86 & 84.00 & 73.17 & 92.77\\
     \hline
     \end{tabular}
  \end{center}
  \caption{Summary of evaluation results}
  \label{result_summary}
\end{table}

%confusion matrix를 먼저 보여주고 그 다음에 train/validation loss 그래프를 보여주자
From the Table 1, our proposed 2sRanking-CNN achieved \textit{Acc} of 96.46\%, \textit{Sp} of 96.00\%, \textit{Se}\textsuperscript{S} of 97.56\%, and \textit{Se}\textsuperscript{G} of 95.18\%. The results are 9.59\% and 10.60\% higher for \textit{Acc} than ranking-CNN and 3-class CNN, respectively. In the case of \textit{Sp}, 2sRanking-CNN was 16\% and 12\% higher than ranking-CNN and 3-class CNN, respectively. Notable is the \textit{Se}\textsuperscript{S} result. The proposed method is 14.63\% higher than ranking-CNN, and especially 24.39\% higher than 3-class CNN. This result shows that ranking-CNN is efficient in the suspicious class where it is relatively continuous and ambiguous in the boundary and that we can obtain more efficient results by introducing 2sRanking-CNN with CAM-extracted ROI method. On the other hand, \textit{Se}\textsuperscript{G} showed more than 90\% accuracy in all three methods and 2sRanking-CNN was only 1.20\% and 2.41\% higher than the other two methods. This result shows that the glaucoma class has distinct characteristics compared to other classes, and conversely, the performance improvement of our 2sRanking-CNN is obtained by efficiently classifying normal and suspicious classes. Similarly, Figure \ref{confusion_matrix} shows that \textit{Sp} and \textit{Se}\textsuperscript{S} are darker in the 2SRanking-CNN than other two methods, which means that the normal and suspicious classes are well classified. As shown in Figure \ref{confusion_matrix}, in ranking-CNN, the rate of misclassification of normal to glaucoma is 4\% while the rate of misclassification of suspicious is 16\%. Likewise, in the 3-class CNN, 5\% and 11\%, respectively. On the other hand, in 2sRanking-CNN, the rate of misclassification of normal to glaucoma is 3\%, which is not much different from that of the two methods, but the ratio of misclassified as suspicious is only 1\%.

\begin{figure}[!tbh]
	\centering
	\includegraphics[width=0.95\textwidth]{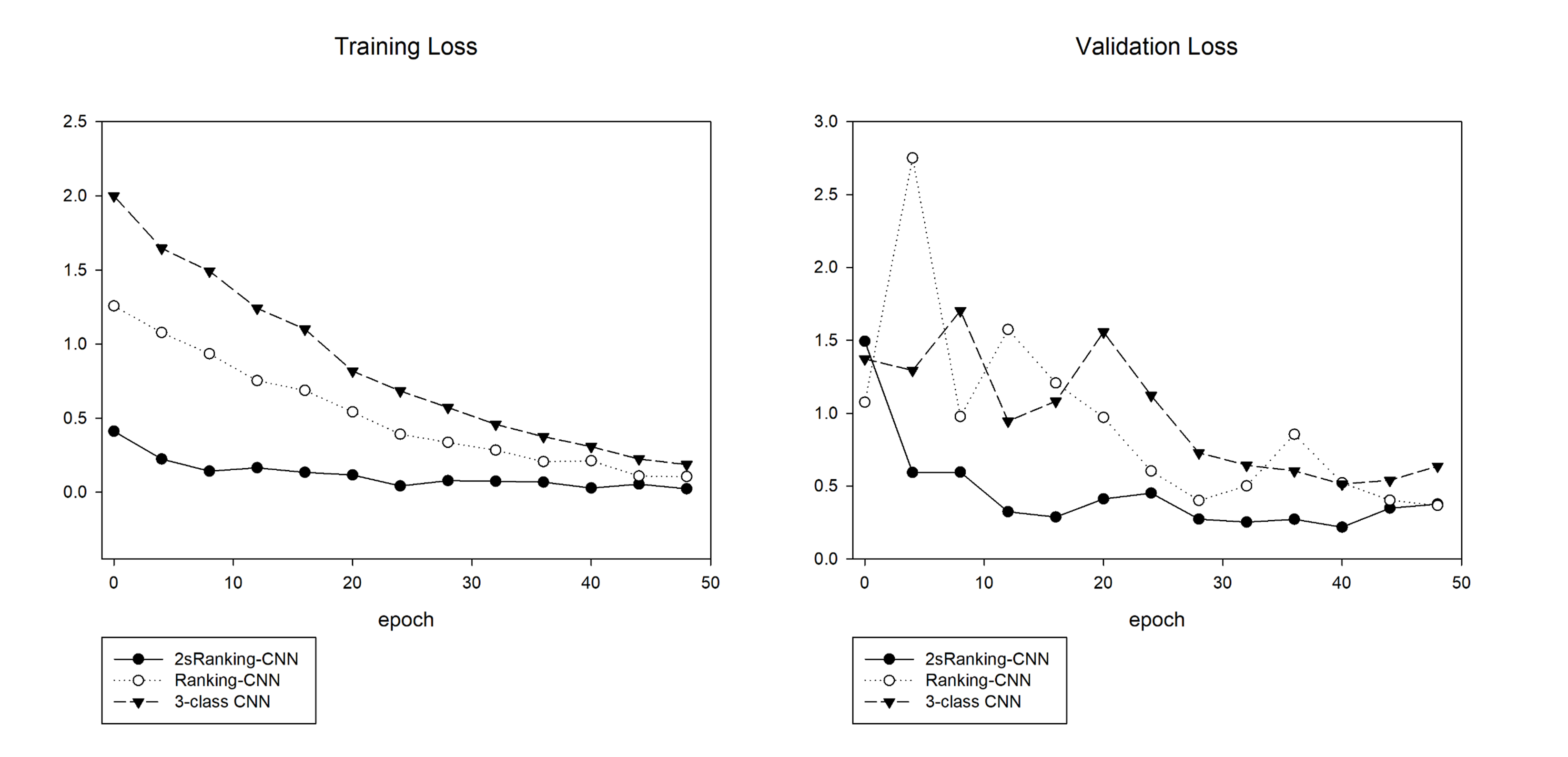}
	\caption{Training and validation loss of 2sRanking-CNN, ranking-CNN, and 3-class CNN}
	\label{loss}
\end{figure}

Figure \ref{loss} shows the training and validation loss for each method. Incidentally, the loss of 2sRanking-CNN means a loss for a total of 50 epochs in 2nd-stage ranking-CNN. From the Figure \ref{loss}, we can observe that 2sRanking-CNN decreases rapidly both in training and validation loss. This is somewhat self-evident, as 2sRanking-CNN allows a second training of the filtered image itself, so that loss can be reduced rapidly and more detailed features can be learned.

\subsection{Comparison with physician's criteria}
\begin{figure}[!tbh]
	\centering
	\includegraphics[width=0.9\textwidth]{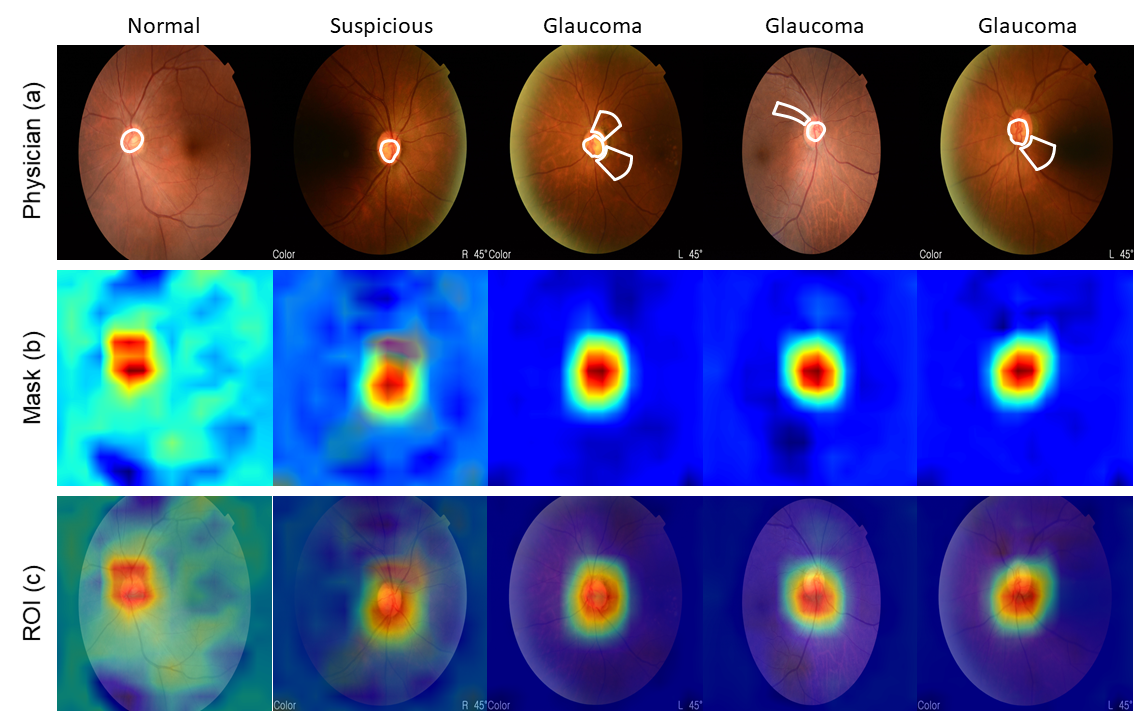}
	\caption{Comparison between diagnostic criteria of the physician and extracted ROI}
	\label{compare_phy}
\end{figure}

%엄영섭 교수님이 주신 판단 근거 이미지와 CAM (또는 CAM + ORIGINAL) 이미지를 같이 보여주자
We attempted to determine how much the mask filter and ROI image extracted by 2sRanking-CNN included glaucoma judgment criteria of ophthalmologists. Figure \ref{compare_phy} shows the comparison between the decision area of the physician, white border in (a), and the mask filter (b) and ROI (c) image from 2sRanking-CNN. From the Figure  \ref{compare_phy} it seems that our ROI image well characterizes the disk/cup of the fundus image. However, the characteristics of the blood vessel around the disk/cup seem to be poorly specified. Nonetheless, the fact that we have high classification accuracy means that there are features inside CNN that we can not fully understand yet.
%-------------------------------------------------------------------------
\section{Discussion}
\label{sec:discussion}
%전체 실험 결과 등을 간략하게 요약하고 어떤 의미를 갖는지 기술해주자
%한계점도 간략하게 써주자 (같은 IOL을 갖는 장비로 촬영한 이미지여서 다른 IOL을 갖는 장비로 촬영하고 
%Transfer-Learning이 가능하다면, 임상적으로 더 의미있는 연구 결과가 될 수 있다는 점을 써주자)
In this paper, we proposed an efficient 2-stage ranking-CNN which classifies normal, suspicious, and glaucoma in the fundus image. We extracted the CAM as a mask filter in the 1st-stage ranking-CNN which trained lightly with train-set and validation-set to efficiently classify the suspicious class which is continuous and ambiguous between normal and glaucoma. The CAM-extracted ROI is used as the input of the 2nd-stage ranking-CNN and the final prediction is obtained through a fully trained process. The results showed that 2sRanking-CNN was average 10.01\%, 14.00\%, 19.51\%, and 1.81\% higher in the \textit{Acc}, \textit{Sp}, \textit{Se}\textsuperscript{S}, and \textit{Se}\textsuperscript{G} than the other two methods. Especially, the accuracy of classifying suspicious class was much higher than the other two methods. When we look at the confusion matrix, we can see that our proposed 2sRanking-CNN well distinguishes between normal and suspicious. On the other hand, we found that the extracted mask and ROI image contain some degree of physician diagnostic criteria.

Despite this excellence, our research also has a limitation. A typical limitation is that our fundus image dataset is 992 total, which is insufficient to fully-train deep CNN. If the number of images is increased, a more general experimental result will be obtained. Thus, we are continuing to add additional patients' fundus images and will have further experiments with more images for future work.
%-------------------------------------------------------------------------
\section{Conclusion}
\label{sec:conclusion}
%전체 연구를 간략하게 요약하고 Future work를 뒤에 기술하자
2-stage ranking-CNN, which classifies normal, suspicious, and glaucoma efficiently from fundus image, has been proposed. Experimental results show that the proposed method efficiently classifies continuous or ambiguous classes compared to existing ranking-CNN and multi-label CNN. In addition, the intermediate result, CAM-extracted ROI, was found to contain some degree of glaucoma judgment criteria of ophthalmologists. The proposed method can be effectively applied to all types of medical image datasets having an intermediate state between normal and diseased states. For the future work, we plan to get more general experimental results by adding a fundus image. In addition, we plan to generalize our 2sRanking-CNN to 2s-CNN which can be applied to multi-label CNN.
%-------------------------------------------------------------------------

\bibliography{egbib}
\end{document}